\documentclass[letterpaper, 10 pt, journal, twoside]{ieeetran}

\usepackage{subcaption}
\usepackage{amsmath} 
\usepackage{amssymb}  
\usepackage{mathtools}
\usepackage{graphicx}
\usepackage[numbers]{natbib}
\usepackage{float}
\usepackage[hyphens]{url}
\usepackage[hidelinks]{hyperref}
\usepackage{booktabs}
\hypersetup{breaklinks=true}

\usepackage{cleveref}
\usepackage{algorithm}
\usepackage[noend]{algpseudocode}
\usepackage{fancyvrb}

\usepackage{xcolor}

\title{
IKFlow: Generating Diverse Inverse Kinematics Solutions
}

\author{Barrett Ames*$^{1}$ \and Jeremy Morgan*$^{2}$ \and George Konidaris$^{3}$
\thanks{Manuscript received: January, 7, 2022; Revised March, 29, 2022; Accepted April, 24, 2022.}%
\thanks{This paper was recommended for publication by Editor Tamim Asfour upon evaluation of the Associate Editor and Reviewers' comments.
This work was supported by DARPA, ONR, and NDSEG}%
\thanks{*, equal contribution}%

\thanks{$^{1}$First author is with Duke University Computer Science Department,
        {\tt\small cbames@cs.duke.edu}}%
\thanks{$^{2}$Second author is with Third Wave Automation,
        {\tt\small jeremy@thirdwave.ai}}%
\thanks{$^{3}$Third authord is with Brown University Computer Science Department,
        {\tt\small gdk@cs.brown.edu}}
        \thanks{Digital Object Identifier (DOI): see top of this page.}
        }%

\begin{document}

\markboth{IEEE Robotics and Automation Letters. Preprint Version. Accepted May, 2022}{Ames \MakeLowercase{\textit{et al.}}: IKFlow} 

\maketitle

\begin{abstract}
Inverse kinematics---finding joint poses that reach a given Cartesian-space end-effector pose---is a fundamental operation in robotics, since goals and waypoints are typically defined in Cartesian space, but robots must be controlled in joint space. However, existing inverse kinematics solvers return a single solution, in contrast, systems with more than 6 degrees of freedom support infinitely many such solutions, which can be useful in the presence of constraints, pose preferences, or obstacles. We introduce a method that uses a deep neural network to learn to generate a diverse set of samples from the solution space of such kinematic chains. The resulting samples can be generated quickly (2000 solutions in under 10ms) and accurately (to within 10 millimeters and 2 degrees of an exact solution) and can be rapidly refined by classical methods if necessary.
\end{abstract}

\section{Introduction}
\IEEEPARstart{I}{nverse}
 Kinematics (IK) maps a task-space Cartesian pose to a joint space configuration, which is a critically important operation for several reasons. For example, it is typically easier to define tasks in a specialized Cartesian coordinate frame that is easily interpretable by the robot's operator---for example, specifying a curve to draw on a whiteboard is more easily done in the frame of the whiteboard  than in the joint space of the robot.  Similarly, it is beneficial to express a grasp pose in Cartesian space because it leaves the robot free to choose among a multitude of valid joint space poses. Of course, each Cartesian-space goal must be translated to a joint pose for the robot to control to; therefore, these scenarios are only possible with a fast IK solver. 

Although there are several open source IK packages~\cite{Beeson2015TRAC-IK:Kinematics,Zhang2017IKBT:Tree, Dai2019GlobalOptimization}, their functionality is still incomplete. Analytical solvers like IKFast~\cite{Diankov2010AutomatedPrograms} and IKBT~\cite{Zhang2017IKBT:Tree} are fast and return all the solutions for an arm, but cannot be applied to arms with more than 6 degrees of freedom (DOF). Numerical solvers can be applied to arms with any number of joints, but only return a single solution, if one is found at all. A solver that can provide many solutions for an arm with 7-DOF or greater, and do so quickly, would add significant functionality that would support more robust robot planning and control. More specifically, it allows for multiple joint solutions to be evaluated for any particular end effector pose. This is important for grasping in cluttered environments because there may be a large number of solutions that are in collision with an object. A large number of solutions is also important for the pathwise-Inverse Kinematics problem \cite{Rakita2019STAMPEDE:Kinematics}. A complete mapping between joint space and task space enables planning to take place in the task space. IK can then be used to validate the task space plan with high confidence. Planning in the task space has a smaller dimension than joint space. This smaller dimension is useful for learning and sampling based approaches because fewer samples are required to cover the space with the same density.

\begin{figure}[t]
\centering
\includegraphics[scale=0.16]{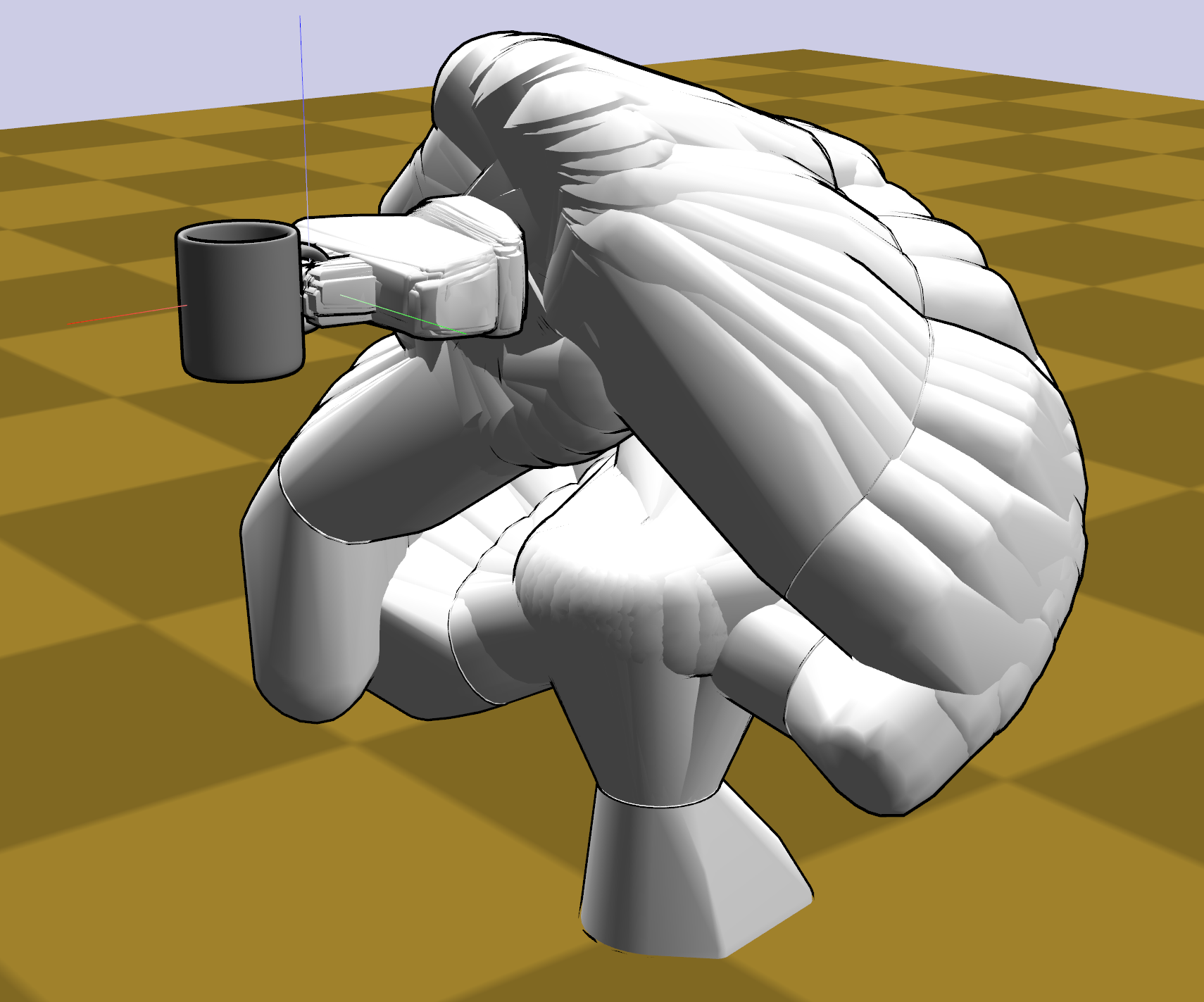}
\caption{One hundred  solutions generated by IKFlow, for a Panda arm reaching to a given end effector pose.}
\label{fig:panda_rej}
\end{figure}
%

An ideal 7+ DOF IK solver should return a set of solutions that covers the entire solution set. These samples should be near-exact solutions. Additionally, it should do so quickly. This allows the solver to serve as a primitive in the inner loop of higher-level decision-making algorithms.  Of these three requirements, accuracy is the least important because verifying a sample using  forward kinematics  is very fast, and  numerical IK  solvers can be seeded with an approximate solution to rapidly refine to any desired accuracy.  

We propose IKFlow, a new IK method that satisfies these requirements by training a neural network to output a diverse set of poses that approximately satisfy a given Cartesian goal pose. By viewing the problem as generative modeling problem over the solution space, we are able to exploit recent advances in Normalizing Flows \cite{Rezende2015VariationalFlows}. Normalizing Flows are a generative modeling approach capable of modeling multi-modal distributions with nonlinear interactions between variables. We show, using several different kinematic models, that IKFlow can be trained---once off, per-robot---to rapidly generate hundreds to thousands of diverse approximate solutions; we also provide a practical open-source implementation of our approach.\footnote{\url{https://sites.google.com/view/ikflow}}


%

%
\begin{figure*}[t]
    \centering
    \includegraphics[scale=.4]{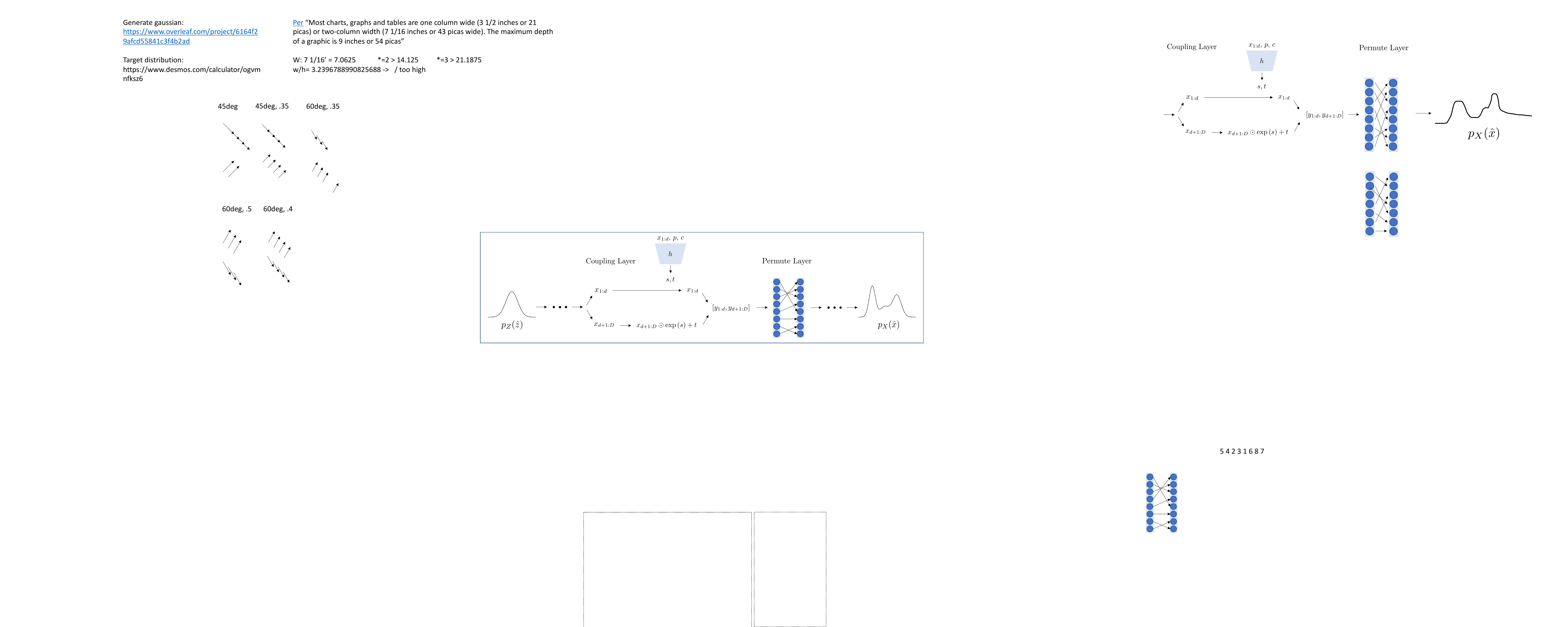}
    \caption{The basic architecture of a conditional normalizing flow network. The base distribution sample is fed into a coupling layer. The coupling layer contains a coefficient network which transforms a subset of the sample. The permutation changes the order of the samples so that the coefficient network of the next coupling layer affects a different subset of the sample. Conditional information is passed into every coupling layer.}
    \label{fig:arch_nf}
\end{figure*}
%

\section{Background}

\subsection{Inverse Kinematics (IK)}
Inverse Kinematics defines the mapping from a robot's operational space to its joint space: 
\begin{equation*}
    f: P \rightarrow  T^n,
\end{equation*}
where $n$ is the number of joints. The topology of the joint space is the n-dimensional torus, $T^n = S^1 \times S^1 ... \times S^1$. The topology of $P$ is determined by the workspace of the robot, but we focus on the case where the end effector pose is in $P \coloneqq SE(3)$. If the robot has joint limits then the topology of the mapping changes: 
\begin{equation*}
    f: P \rightarrow R^n,
\end{equation*}
where $R^n$ is the n-dimensional Euclidean space. This topological difference is important because it allows for many robots to be modeled by generic density estimators, instead of special purpose estimators built for tori~\cite{Rezende2020NormalizingSpheres}. 
\subsubsection{IK for 7+ DOF}
When the degrees of freedom of the robot exceed the degrees of freedom of the operational space (i.e. the robot is kinematically redundant) there are infinitely many solutions. This occurs because the extra degree of freedom allows for a continuum of configurations that satisfy the desired Cartesian goal. Thus for a specific pose $p \in SE(3)$ an IK solver should return a subset of joint space:
\begin{equation*}
    f: p \rightarrow R^n_g \subseteq R^n. 
\end{equation*}
There are two current approaches for describing the solution space of 7+ DoF arms. The first category uses only a single point to describe the solution space, but returns quickly, on the order of $1$ millisecond. The second approach returns a more thorough set of solution points, by running the first approach with randomly sampled start states. In order to obtain a representative set of the solution space this approach requires extensive random sampling, and thus is significantly slower, on the order of seconds for thousands of solutions. The approach we detail in Section~\ref{sect:LIK} uses generative modeling to generate upwards of a thousand solutions in under $10$ milliseconds.

\subsection{Deep Generative Modeling}
Generative modeling represents arbitrary distributions in such a way that they can be sampled. This is achieved by transforming known base distributions into target distributions. For example:
\begin{equation*}
    g: N(\hat{0},I) \rightarrow Q,
\end{equation*}
where $Q$ is an arbitrary distribution and $g$ is a neural network. We additionally define the \textit{latent space} as the space in which the base distribution lies - in our case $R^D$, where $D$ is the dimension of the network. We draw samples from the arbitrary distribution by drawing samples from the base distribution and passing them through $g$. There are several different generative modeling methods. In order to select an approach we used the following criteria. First, the diversity of samples returned is important to ensure that the full breadth of the solution space is returned. Second, the approach must be able to handle multi-modal data and nonlinear dependencies between variables in order to cover the solution space. Third, the method must be capable of handling conditional information, because the solution space is conditioned on the Cartesian goal. Fourth, the method must sample solutions quickly because IK is often used as a primitive by other procedures. Finally, the sampling procedure must produce samples with sufficient accuracy. Given these desired properties we propose to use a normalizing flow approach.

%
\begin{table*}[t]
\centering
\begin{tabular}{ lcccc }
\toprule
Robot & DOF  & Coupling Layer Width & Coupling Layers & Number of Parameters \\ 
\midrule
ATLAS (2013) - Arm and Waist &  9 &  15 &  9 & $3.836 \times 10^7$ \\
ATLAS (2013) - Arm &  6 &  15 &  12 &  $5.115 \times 10^7$  \\
Baxter &  7 &  15 &  16 & $6.820 \times 10^7$   \\
Panda &  7 &  9 &  12 & $5.093 \times 10^7$   \\
PR2 &  8 &  15 &  8 & $3.410 \times 10^7$   \\
Robonaut 2 - Arm and Waist &  8 &  15 &  12  & $5.115 \times 10^7$    \\
Robonaut 2 - Arm &  7 &  15 &  12  & $5.115 \times 10^7$  \\
Valkyrie - Whole Arm and Waist &  10 &  15 &  16  & $6.820 \times 10^7$   \\
Valkyrie - Lower Arm &  4 &  15 &  9 & $3.836 \times 10^7$  \\
Valkyrie - Whole Arm &  7 &  15 &  12  & $5.115 \times 10^7$ \\
\bottomrule


\end{tabular}
\caption{The primary parameter to select when fitting a network to a new kinematic chain is the number of coupling layers. The number of coupling layers increases the expressivity of the invertible portion of the network. The network can also be made more expressive by increasing the coupling layer width. When the width becomes larger than the DOF the system becomes more capable of representing multi-modal target distributions.  }
\label{table:parameters}
\end{table*}
%

\subsubsection{Normalizing Flows}
Normalizing Flows are a generative modeling approach that provides quick sampling, stable training and arbitrary data distribution fitting~\cite{Papamakarios2019NormalizingInference}. Normalizing flows are based on two components. The first component is a series of functions that are easy to invert. These easily invertible functions enable the same network to quickly estimate densities and produce samples. In the literature, this series of functions is commonly referred to as the \textit{coupling layers}. Sampling is performed by passing a sample from the base distribution through each of the coupling layers: 
\begin{equation}
    \hat{x} = f_1 \circ f_2 ... \circ f_n(\hat{z}), \hat{z} \sim N(\hat{0},I),
    \label{eq:norm_flow}
\end{equation}
where $\hat{x}$ is the sample in the data space and $\hat{z}$ is a sample from the base distribution. The second important component is the efficient calculation of the Jacobian's determinant. This enables density estimation of a data point to be computed with the change of variables formula: 
\begin{equation*}
    \label{eq:change_var}
  p_{X}(\hat{x})=p_{Z}(\hat{z})\left|\operatorname{det}\left(\frac{\partial g(\hat{z})}{\partial \hat{z}^{T}}\right)\right|^{-1}.  
\end{equation*}
The change of variables formula allows a distribution over one set of variables to be described by another set of variables given the determinant of the Jacobian between the two variables. A normalizing flow uses this along with a simple prior distribution, $p(z)$---here a Normal distribution---to enable density estimation of the data distribution $p(x)$. To make the determinant of the Jacobian tractable,  special coupling layers are used. The coupling layer used in IKFlow was developed by~\citet{Kingma2018Glow:Convolutions}: 
\begin{align}
y_{1: d} &=x_{1: d} \\
s,t &= h(x_{1:d}) \\ 
y_{d+1: D} &=x_{d+1: D} \odot \exp \left(s\right)+t,
\label{eq:inv_layer}
\end{align} where $x$ is the input data to a layer, and $y$ is the output. $s$ affects the scaling of the layer, $t$ shifts the input of the layer, and $d = \left \lfloor{D / 2}\right \rfloor $. This layer is then inverted by: 
\begin{align*}
x_{1: d} & =y_{1: d} \\
s,t &= h(x_{1:d}) \\ 
x_{d+1: D} & =\left(y_{d+1: D}-t\right) \odot \exp \left(-s\right).
\end{align*}
An important property of the layers is that they can be inverted without inverting the function $h$ that produces $s$ and $t$. $h$, also known as the coefficient network, can therefore be arbitrarily complex. Further, by holding $\hat{x}_{1:d}$ constant through the transformation, the Jacobian will have zeros in the upper diagonal, and thus the determinant will only be the product of the diagonals of the Jacobian.

\subsubsection{Conditional Normalizing Flows}
Conditional Normalizing Flows change the scaling and shifting of a coupling layer based on conditional information. The conditional information, $\kappa$, is passed as part of the input to the network which estimates $s$ and $t$. Now $h$ is a function of both $x_{1:d}$ and $\kappa$. The invertibility of the coupling layer remains unaffected because the conditioning information is passed into $h$ which need not be inverted to invert the layer. Figure~\ref{fig:arch_nf} provides a simple diagram of the architecture. While other methods have been proposed for conditioning normalizing flows (i.e. ~\citet{Ardizzone2019AnalyzingNetworks}) this formulation reduces the number of hyperparameters that must be tuned, and has a simple Maximum Likelihood training loss: 
\begin{equation}
\label{eq:MLE}
    \mathcal{L} = -\log(p_{Z}(z)) - \log \left|\operatorname{det}\left(\frac{\partial g(z|\kappa)}{\partial z^{T}}\right)\right|^{-1}. 
\end{equation}

For a more thorough review of Normalizing Flows see ~\citet{Papamakarios2019NormalizingInference}. 

\subsubsection{Maximum Mean Discrepancy (MMD)}
Finally, a method for comparing distributions is necessary to evaluate the performance of a generative model. We use Maximum Mean Discrepancy (MMD)  because of its strong theoretical properties and ease of implementation. Given two distributions, $P(X)$ and $Q(Y)$, MMD computes the distance between the two distributions as the squared distance between the mean of the embeddings of the distributions. 
\begin{equation*}
    \text{MMD}(P,Q) = || E_{X\sim P}[\phi(X)] - E_{Y \sim. Q}[\phi(Y)] ||. 
\end{equation*}
Notably, the embeddings must be in a Reproducing Kernel Hilbert Space (RKHS). The capability of MMD to capture the distance between two distributions is dependent on the selection of a good embedding function $\phi$. We selected the inverse multi-quadric kernel because of its long tails and previous use~\citet{Ardizzone2019AnalyzingNetworks}. A more detailed analysis of Maximum Mean Discrepancy is provided by~\citet{Gretton2012ATest}. 

%
\begin{figure*}
     \centering
     \begin{subfigure}[b]{0.22\textwidth}
         \centering
         \includegraphics[width=\textwidth]{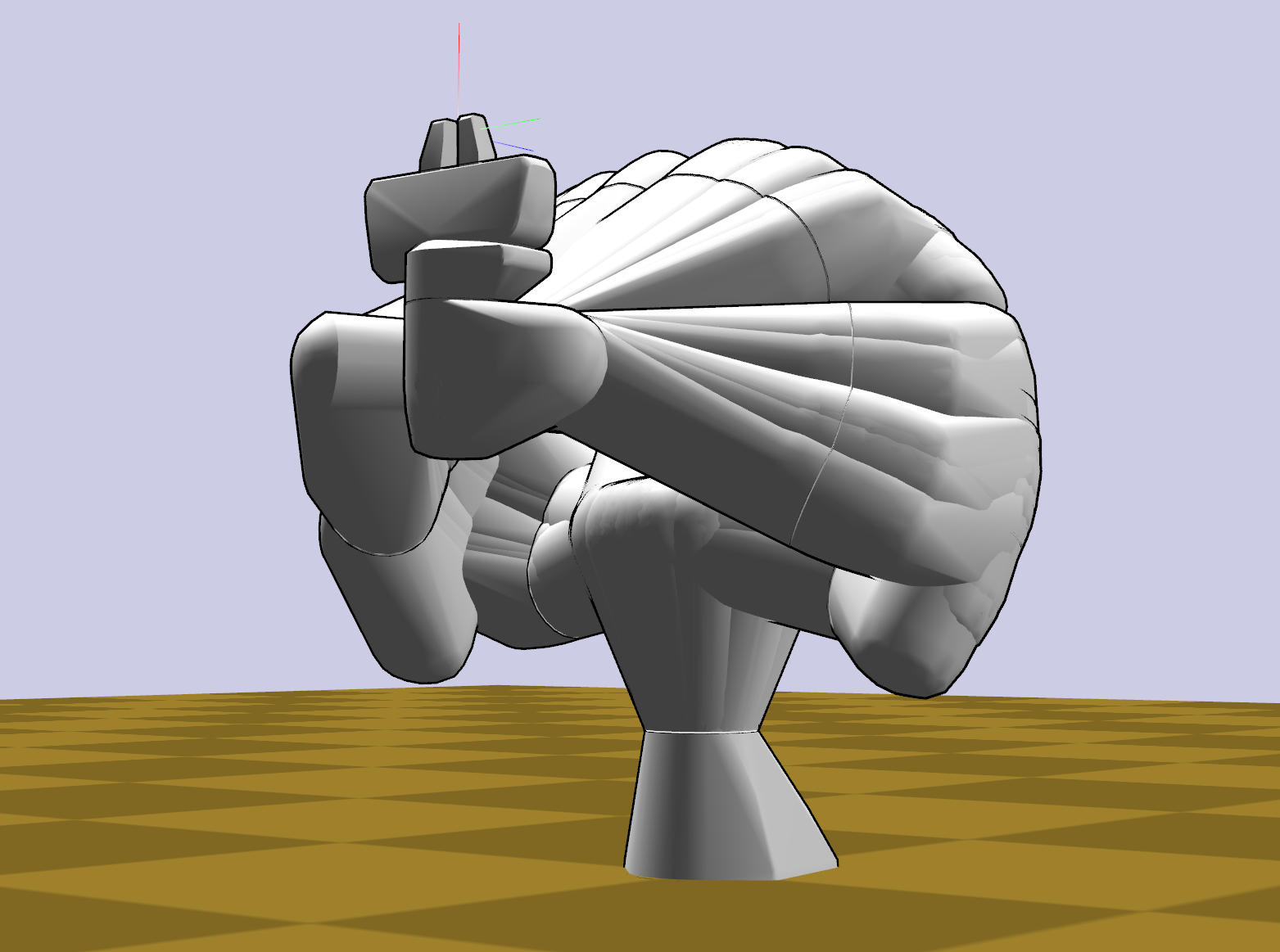}
         \caption{Panda sampled solutions}
         \label{fig:panda_gt}
     \end{subfigure}
     \hfill
     \begin{subfigure}[b]{0.22\textwidth}
         \centering
         \includegraphics[width=\textwidth]{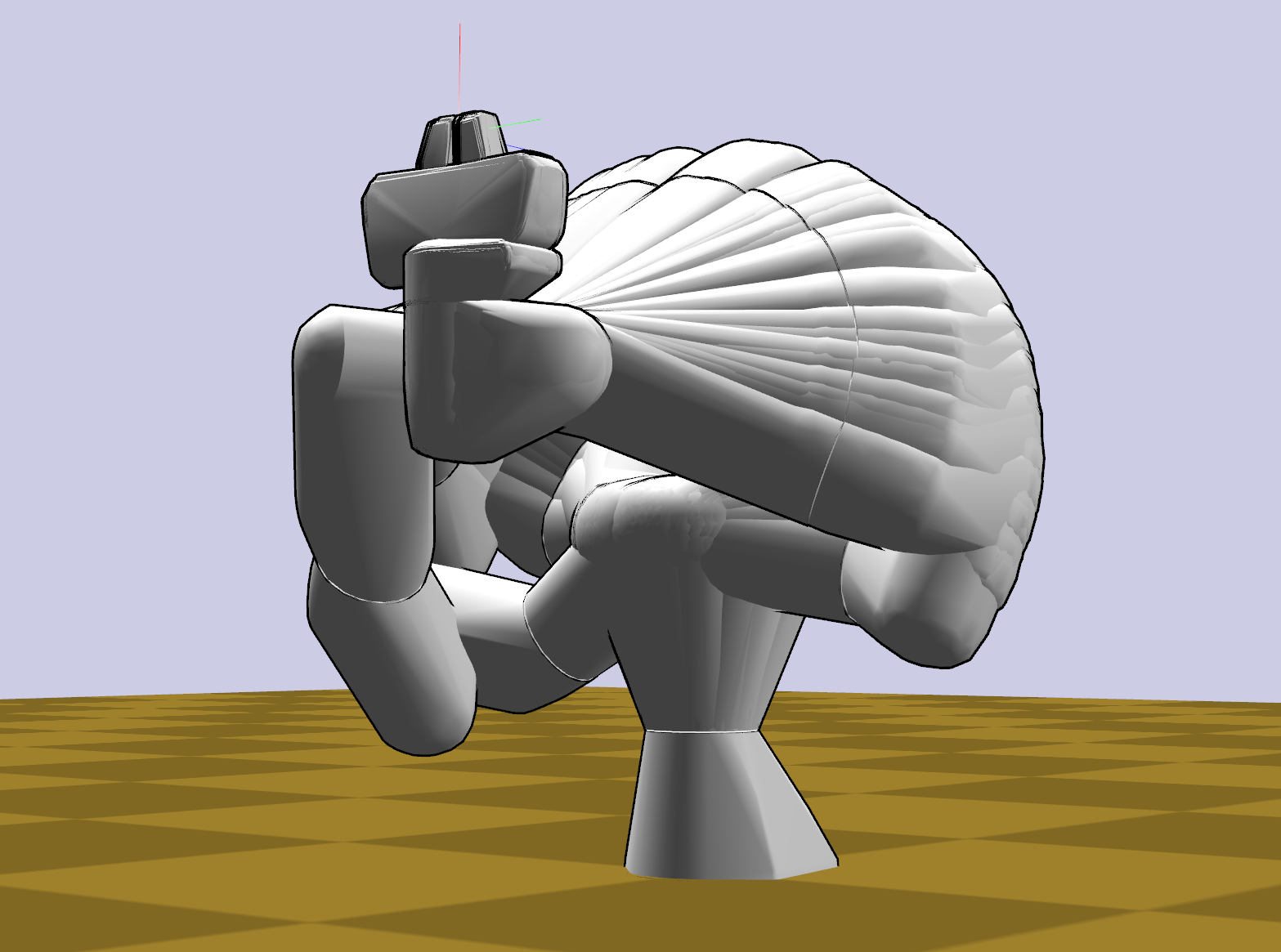}
         \caption{Panda IKFlow solutions}
         \label{fig:panda_ikflow}
     \end{subfigure}
     \hfill
     \begin{subfigure}[b]{0.22\textwidth}
         \centering
         \includegraphics[width=\textwidth]{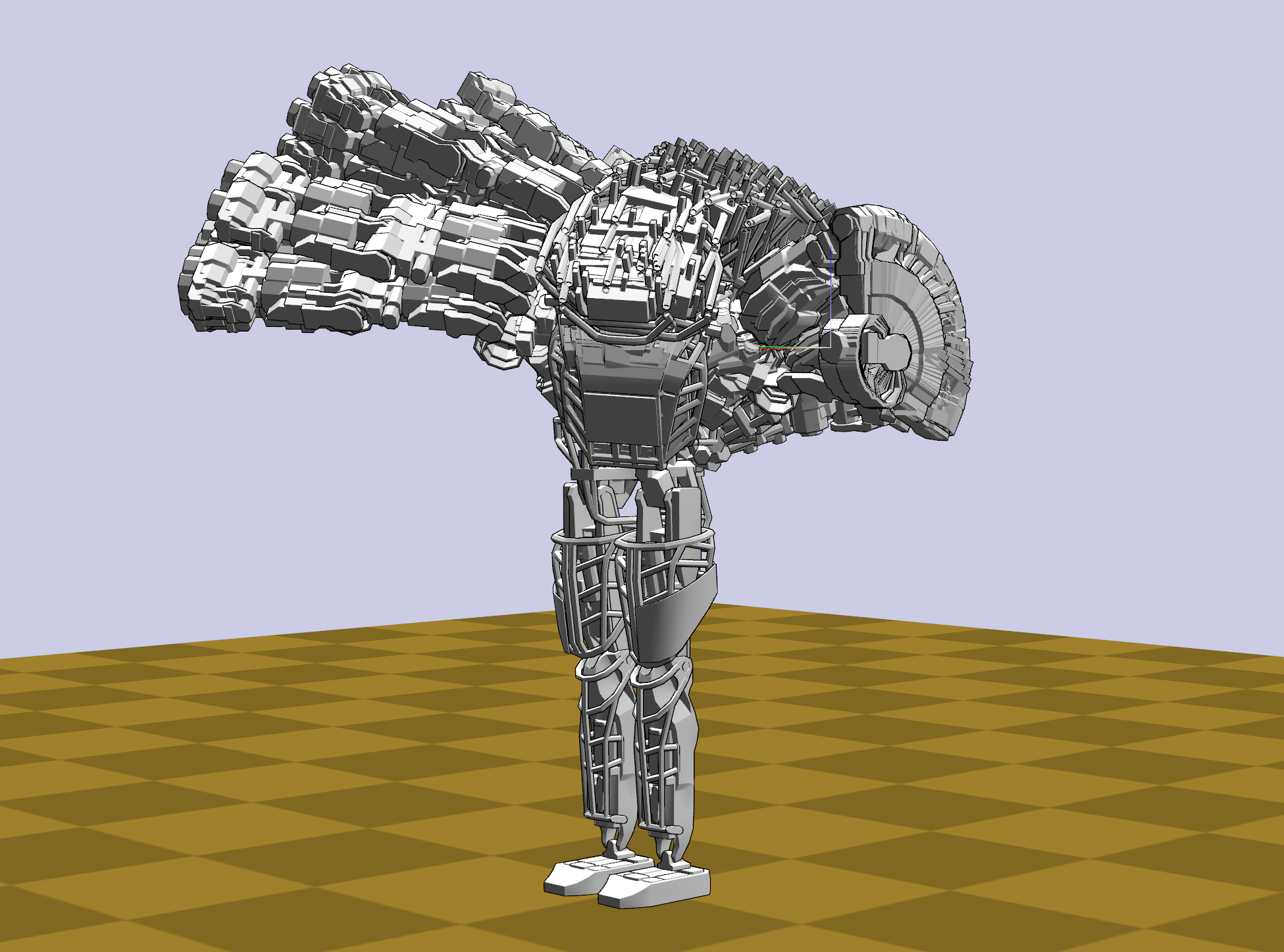}
         \caption{ATLAS sampled solutions}
         \label{fig:atlas_gt}
     \end{subfigure}
     \hfill
     \begin{subfigure}[b]{0.22\textwidth}
         \centering
         \includegraphics[width=\textwidth]{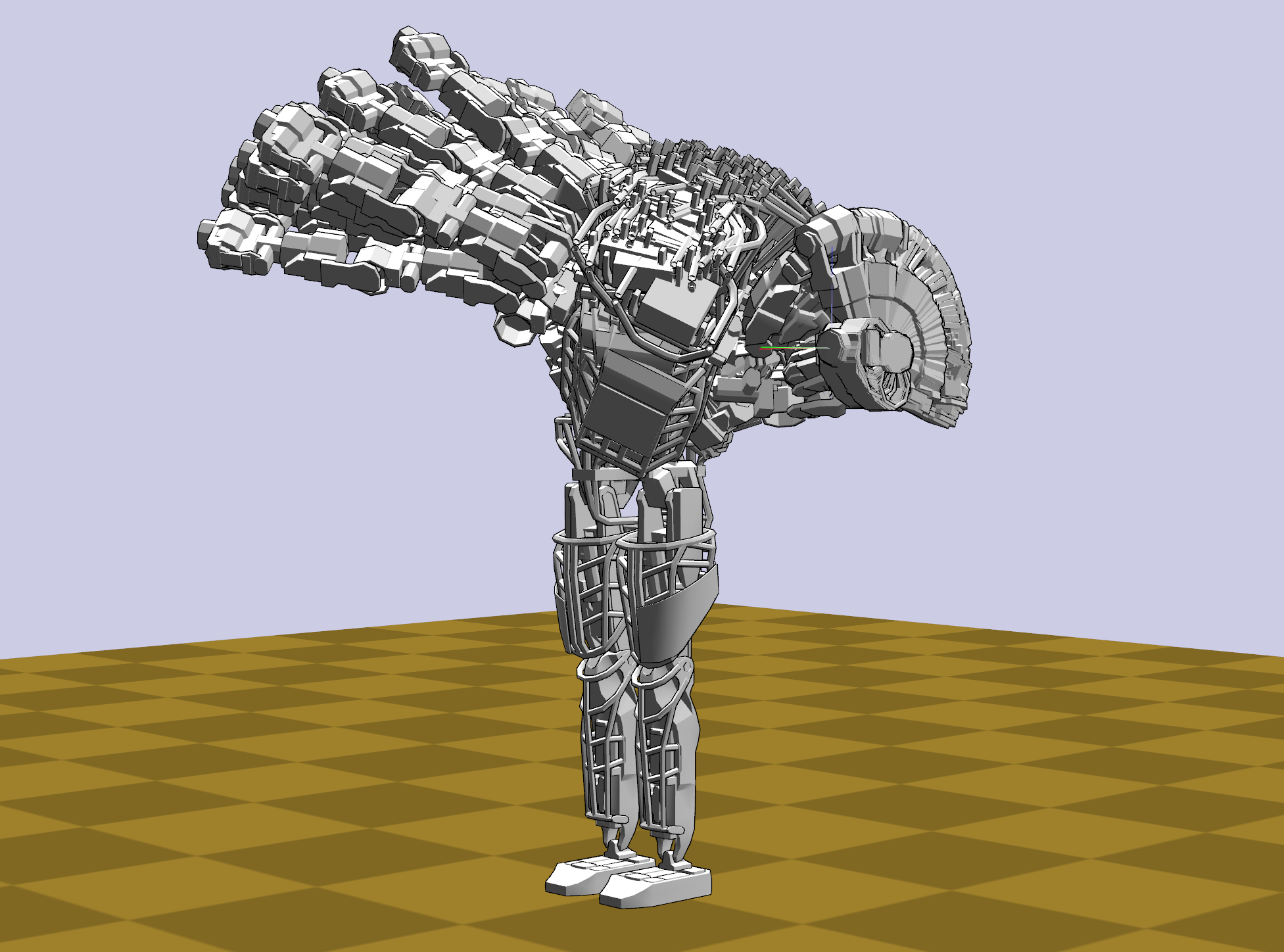}
         \caption{ATLAS IKFlow solutions}
         \label{fig:atlas_ikflow}
     \end{subfigure}
        \caption{A comparison of samples created by providing random seeds to TRAC-IK and samples from IKFlow for two robots, the Panda arm and ATLAS - Arm and Waist. Qualitatively, the solutions returned by the IKFlow models look to cover the same space of solutions found through rejection sampling. The MMD scores for the Panda and ATLAS solutions are respectively 0.0122 and 0.0081. A lower MMD score implies distributions which are more similar.}
        \label{fig:comparisons}
\end{figure*}
%

\section{Related Work}
There have been a number of attempts to use neural networks for inverse kinematics~\cite{Almusawi2016AVP6242, Csiszar2017OnNetworks, DembyS2019ANetwork, Ren2020LearningNetworks}. Most of these approaches do not solely implement the inverse kinematics function. For example, \citet{Csiszar2017OnNetworks} incorporate the problem of calibrating the physical robot with its internal model. \citet{Almusawi2016AVP6242} learn the kinematics model in addition to performing Cartesian control. \citet{DembyS2019ANetwork} create a neural network approach that focuses exclusively on the IK problem, but conclude that it is not a fruitful path because of large error.

The architectures used in the previous approaches are all fundamentally limited as they can only return a single solution for a particular input. However, IK is a one-to-many mapping, and access to additional solutions may prove useful for the control layer above IK. \citet{Ren2020LearningNetworks} take a generative approach by using several different types of GANs, but the error of samples is worse than a fully connected network, with the best performing approach achieving 8cm of error. \citet{Ardizzone2019AnalyzingNetworks} apply a similar deep generative approach to IKFlow but only to a small planar IK problem. We extend and refine their work in three key ways. First, we decrease the amount of error present in the solutions by adopting a conditional Invertible Neural Network. Second, we improve training stability by addressing a corner case related to the dimensionality of solution spaces. Third, we expand the coverage to include the entire physical workspace of the arm. 

\citet{Kim2021LearningManipulator} are the most similar point of comparison because they also use normalizing flows to solve inverse kinematics. IKFlow has several differences, the use of Conditional Normalizing Flows, the addition of training noise, and a sub-sampling procedure for the base distribution. These differences result in better performance for theoretical and practical reasons that are explained in the following section.  Additionally, IKFlow doesn't have a network that can calculate the forward kinematics, because the forward kinematics of a system are easy to calculate analytically. IKFlow's single network can perform both inverse kinematics and density estimation, which are difficult to perform analytically but with reduced overall complexity. As a result of these differences, the accuracy achieved by IKFlow is $\sim5$x better in position, and $\sim3$x better in orientation, as measured with the Panda Arm and compared with Figure 3 in \citet{Kim2021LearningManipulator}.


%

\section{IKFlow: Learned Inverse Kinematics}
\label{sect:LIK}
The first step to learning inverse kinematics is to treat the solution space as a distribution that has a uniform probability, which may have multiple intervals, and is conditioned on the desired pose. We model the solution distribution with Conditional Normalizing Flows because they are quicker than alternatives to sample, are stable when training, and are capable of representing the full solution space~\cite{Kobyzev2021NormalizingMethods}. Additionally, they don't face problems such as vanishing gradients or mode collapse which are common when training Generative Adversarial Networks (GANs), a different neural generative modeling approach \cite{Kobyzev2021NormalizingMethods}.

 

There are three steps to applying Conditional Normalizing Flows to the inverse kinematics problem. First, a data set must be constructed. Second, we must select an architecture and parameters of the architecture. Third, we must design a loss function for minimizing the difference between samples from the generative model and the data distribution.

%
\begin{table*}[t]
%
\centering
\begin{tabular}{ lccccc } 

\toprule
    Robot & DOF & MMD Score  & Time (msec) & Average L2 Position Error (millimeters) & Avg Angular Error (degrees) \\
\midrule
ATLAS (2013) - Arm and Waist &  9 &  0.03723 &  4.8 &  2.66 &  0.61 \\
ATLAS (2013) - Arm &  6 &  0.004828 &  6.4 &  1.19 &  0.28 \\
Baxter &  7 &  0.0331 &  8.29 &  4.5 &  1.19 \\
Panda &  7 &  0.0306 &  6.28 &  7.72 &  2.81 \\
PR2 &  8 &  0.2128 &  4.31 &  3.3 &  1.56 \\
Robonaut 2 - Arm and Waist &  8 &  0.03691 &  6.44 &  3.63 &  0.77 \\
Robonaut 2 - Arm &  7 &  0.0327 &  6.45 &  1.91 &  0.46 \\
Valkyrie - Whole Arm and Waist &  10 &  0.0361 &  8.55 &  3.49 &  0.74 \\
Valkyrie - Lower Arm &  4 &  1.508e-06 &  4.87 &  0.36 &  0.15 \\
Valkyrie - Whole Arm &  7 &  0.0292 &  6.45 &  1.64 &  0.38 \\
\bottomrule

\end{tabular}
\caption{ This table contains the results of the experimental runs. Since the IKFlow distribution is compared with a ground truth distribution the MMD score provides a measure of how well the solution distribution from the network covers the whole solution space. Time is the total time to return 100 solutions for one Cartesian pose measured in milliseconds. Average position error and Average angular error are the position and geodesic distance, respectively, between the desired goal pose and the pose achieved from joint solutions.}
\label{table:results}
\end{table*}
%

%
\begin{figure}
    \centering
    \includegraphics[scale=0.5]{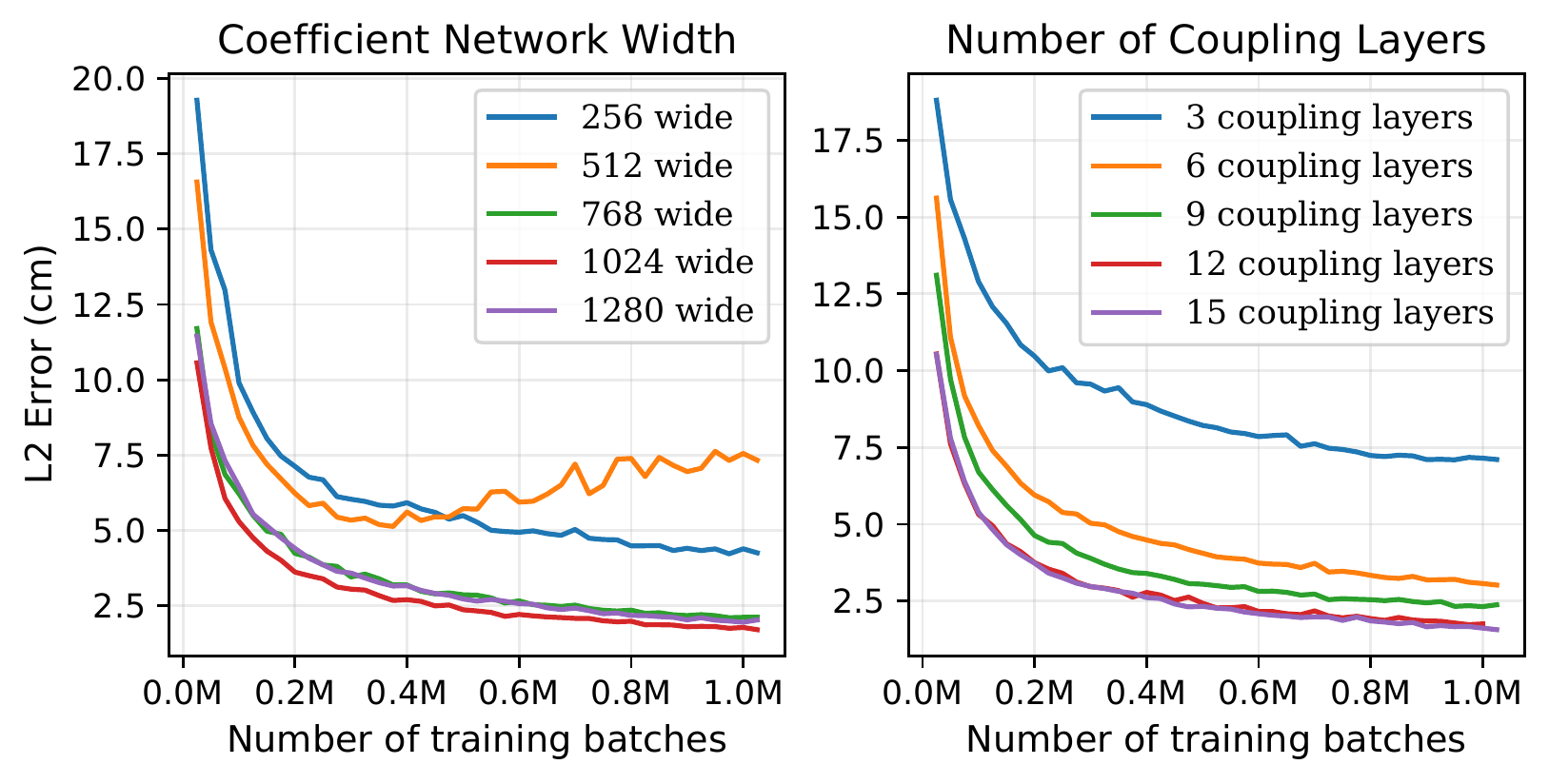}
    \caption{Learning curves for networks with increasing coefficient network widths (left) and number of coupling layers (right) for the Panda Arm robot. For the coefficient network width comparison, the network has 12 coupling layers, where each coefficient network has 3 fully connected layers. A width of 1024 has the highest performance. For the coupling layer comparison, the coefficient function are 3x1024-wide fully connected neural networks with Leaky-Relu activation. }
    \label{fig:ablation}
\end{figure}
%

\subsection{Data Generation}
One advantage of the inverse kinematics problem is the relative ease with which data can be generated. All serial kinematic chains have known forward kinematics that can be used to generate training data. Data is generated by uniformly sampling from the interval defined by the joint limits. The joint samples are then fed through a forward kinematics function to obtain the corresponding Cartesian pose. The Cartesian pose becomes the conditional input to the network and the joint data is used as the target distribution. More sophisticated sampling methods could be used to account for effects at joint limits and for self-collisions, but we were able to obtain satisfactory performance without them and thus leave this investigation for future work.

\subsection{Normalizing Flow Architecture}
The fundamental architecture of the network is defined by the use of normalizing flows, and was detailed in equations~\ref{eq:norm_flow}-\ref{eq:inv_layer}. The remaining design choices include the selection of a base distribution, selecting the number of coupling layers, the width of each coupling layer, and specification of the coefficient network. 

The base distribution affects the complexity of the density estimation and the sampling speed. We chose the Normal distribution because it simplifies the calculation of the Maximum Likelihood Loss function and is quick to sample. The remaining choices for the architecture affect the capability of the network to fit a target distribution. The width of each coupling layer must be at least as large as the degrees of freedom; if it is larger than the DOF it allows for multi-modal distributions to be more easily modeled. The number of coupling layers required depends on the interactions of joints with each other. The more complex the inter-joint relationship, the more coupling layers are required, as more layers allow for more interaction between joints because of the permutation layers. The number of coupling layers and the width of each can be found automatically by performing a hyper parameter sweep over a reasonable range (i.e. [$1$, $30$] and [$n$, $n+10$]).
Finally, the coefficient network must be expressive enough to capture dependencies between the conditional information (i.e. the goal pose) and a subset of the layer values. Table~\ref{table:parameters} details the variable parameters (selected by hyperparameter search) for each kinematic chain that we test. An analysis was performed to demonstrate the impact of the number of coupling layers and the width of the coefficient function networks on the average Positional error of a network Figure ~\ref{fig:ablation}. This demonstrates that the error of a network decreases with increased size up to a point, at which point increasing the network size does not improve performance.


\subsection{Loss Functions}
The Maximum Likelihood loss detailed in equation~\ref{eq:MLE} is theoretically the only loss function necessary for achieving high performance. However, for several of the kinematic chains tested, training diverged when trained with the Maximum Likelihood loss. The cause of this divergence was a mismatch between the dimension of the solution manifolds and the base distribution.

\subsubsection{Solution Sub-manifolds}
In Figure~\ref{fig:panda_gt} the last joint of the arm appears in only one position. This implies that the solution space of that particular pose is not the same dimension as the joint space, because only 6 of the 7 joints in the arm can vary. This is not a unique situation, but occurs throughout the configuration space as some Cartesian poses can only be reached by fixing one of the joints at a specific configuration, for example at one of its limits. This is problematic because the normalizing flow approach does not perform well on distributions that are lower dimension than the base distribution. Specifically, the change of variables equation (\ref{eq:change_var}) does not hold if the base distribution and the target distribution are not the same dimension. One way to ensure the target distribution and the base distribution are the same dimension is to add noise that is the full dimension \cite{Kim2020SoftFlow:Manifolds}. If the magnitude of the added noise is passed in as a conditional variable, its effect can be removed at test time by setting that piece of the conditional to 0: \begin{align*}
        c \sim U(0,1) \\ 
        v \sim N(0,c) \\ 
        \hat{x} = x + v,
\end{align*}
where $c$ and $v$ are drawn every training iteration, and added to the training data. The resulting loss function is: 
\begin{equation}
\label{eq:submanifold_MLE}
    \mathcal{L} = -\log(p_{Z}(z)) - \log \left|\operatorname{det}\left(\frac{\partial g(z|p,c)}{\partial z^{T}}\right)\right|^{-1}, 
\end{equation}
where $p$ is the desired Cartesian pose. 

\subsection{Base Distribution Sub-sampling}

One down side of using the Normal distribution for the base distribution is the tails of the distribution. At evaluation time a point sampled from the tail of the Normal distribution is likely to produce a solution that has high error because the loss function encourages known solutions to be closer to the mean of the distribution. In order to reduce the impact of the tails the base distribution is sub-sampled at test time. This reduces the likelihood of a point from the tail of a distribution, however this encourages less diverse solutions. This trade off is demonstrated by Figure~\ref{fig:latent_scaling}. Thus the scaling of the base distribution should be treated as a tuning parameter for the particular application of IKFlow.  
%

\section{Experiments} 
The models were built with the FrEIA framework~\cite{Ardizzone2019AnalyzingNetworks} and trained with PyTorch~\cite{Paszke2019PyTorch:Library}. Additional details about the network architectures and training parameters can be found in the appendix. 
Once trained the model was evaluated on the three desired quantities: speed, accuracy, and solution space coverage. 

While evaluating a model,  we use a scaling factor of $0.25$, which provides an adequate trade-off between accuracy and diversity. We found that in general, this scaling factor lowers the average positional error by $\sim30$\%, at the expense of an $\sim150$\% increase in MMD Score.

Speed was measured by the time required to sample $100$ solutions for a given Cartesian pose, averaged over $50$ randomly-sampled Cartesian poses. In addition, to understand how the model scales, runtime was measured as the number of requested solutions increased. 

Model accuracy was measured by sampling $1000$ test Cartesian poses and obtaining $250$ joint solutions for each Cartesian pose. The joint solutions were then passed into a forward kinematics routine to compute the realized Cartesian poses. The averaged difference between the $250000$ desired and realized poses was recorded and reported in terms of position and geodesic distance. 

Sample diversity and solution space coverage was measured by calculating the MMD score for ground truth samples and IKFlow samples. The MMD Score is found by taking the average of 2500 Maximum Mean Discrepancy values, each calculated between ground truth samples and the samples returned IKFlow.  For each calculation, 50 joint space solutions are generated by each method for a randomly drawn pose. The minimum possible MMD value is 0. Values close to zero imply that the distributions are more similar; thus when the IKFlow achieves a low MMD score, it provides good coverage of the solution space. The ground truth samples were calculated by providing different seeds to TRAC-IK for the same Cartesian pose.

%
\begin{figure}
    \centering
    \includegraphics[scale=0.45]{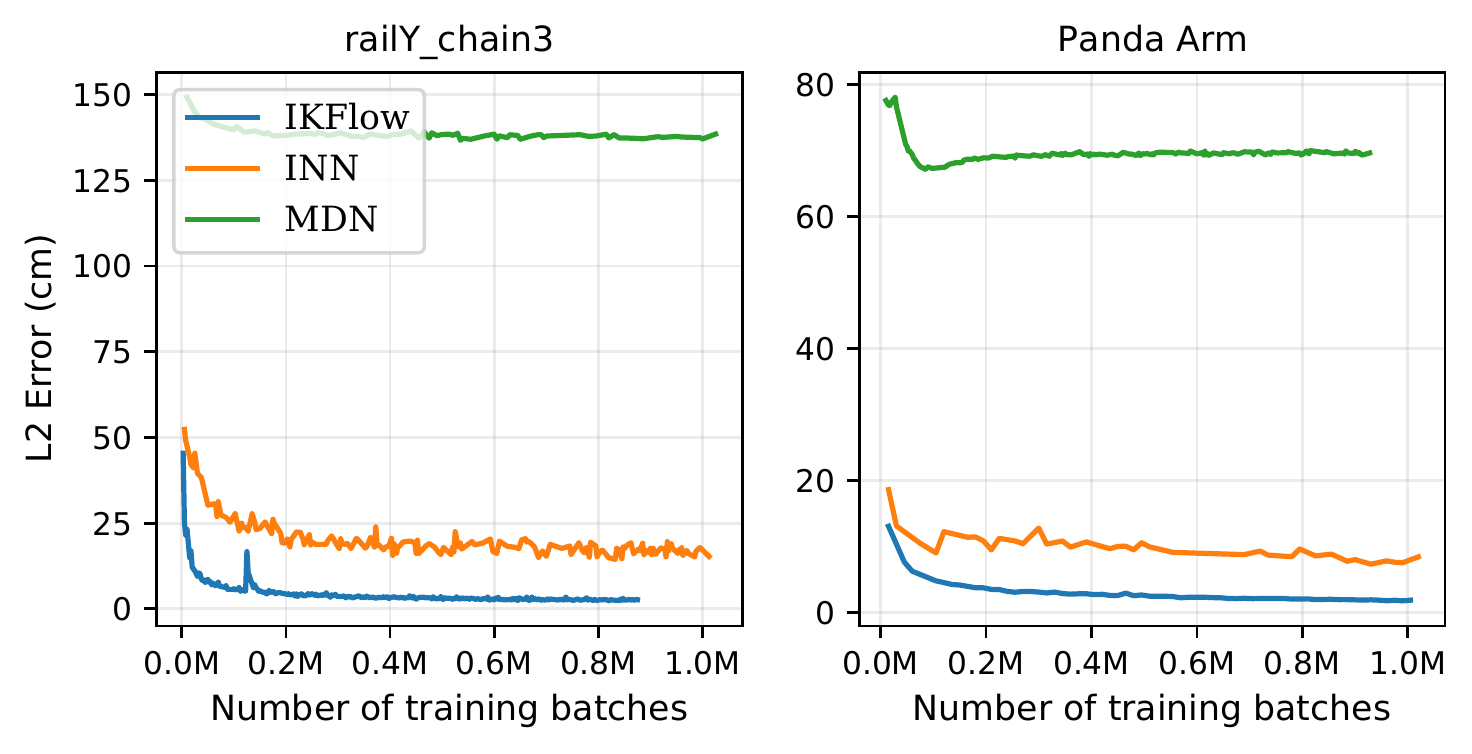}
    \caption{Learning curves for IKFlow, INN, and MDN for the railY\textunderscore chain3 and Panda Arm robots.}
    \label{fig:benchmarking}
\end{figure}
%

In addition, we compare IKFlow to an Invertible Neural Network (INN)  ~\cite{Ardizzone2019AnalyzingNetworks} and to a Mixture Density Network (MDN) ~\cite{Bishop1994MixtureNetworks} on two benchmarks---railY\textunderscore chain3 and Panda Arm. The railY\textunderscore chain3 robot (first used by ~\citeauthor{Ardizzone2019AnalyzingNetworks}), is a planar robot with 4 joints---the first is a prismatic actuator along the y axis with limits [-1, 1]. The final three are revolute joints with limits [$-\pi$, $\pi$] and associated arm segments of length 1. For a fair comparison, the IKFlow and INN models have the same number of coupling layers and size of coefficient networks. Further details about the testing procedure are presented in the appendix. 

These experiments were carried out on $10$ different kinematic chains across $6$ different robots, to evaluate the generality of the approach across different kinematic structures.

%

%
\begin{figure}
    \centering
    \includegraphics[scale=0.45]{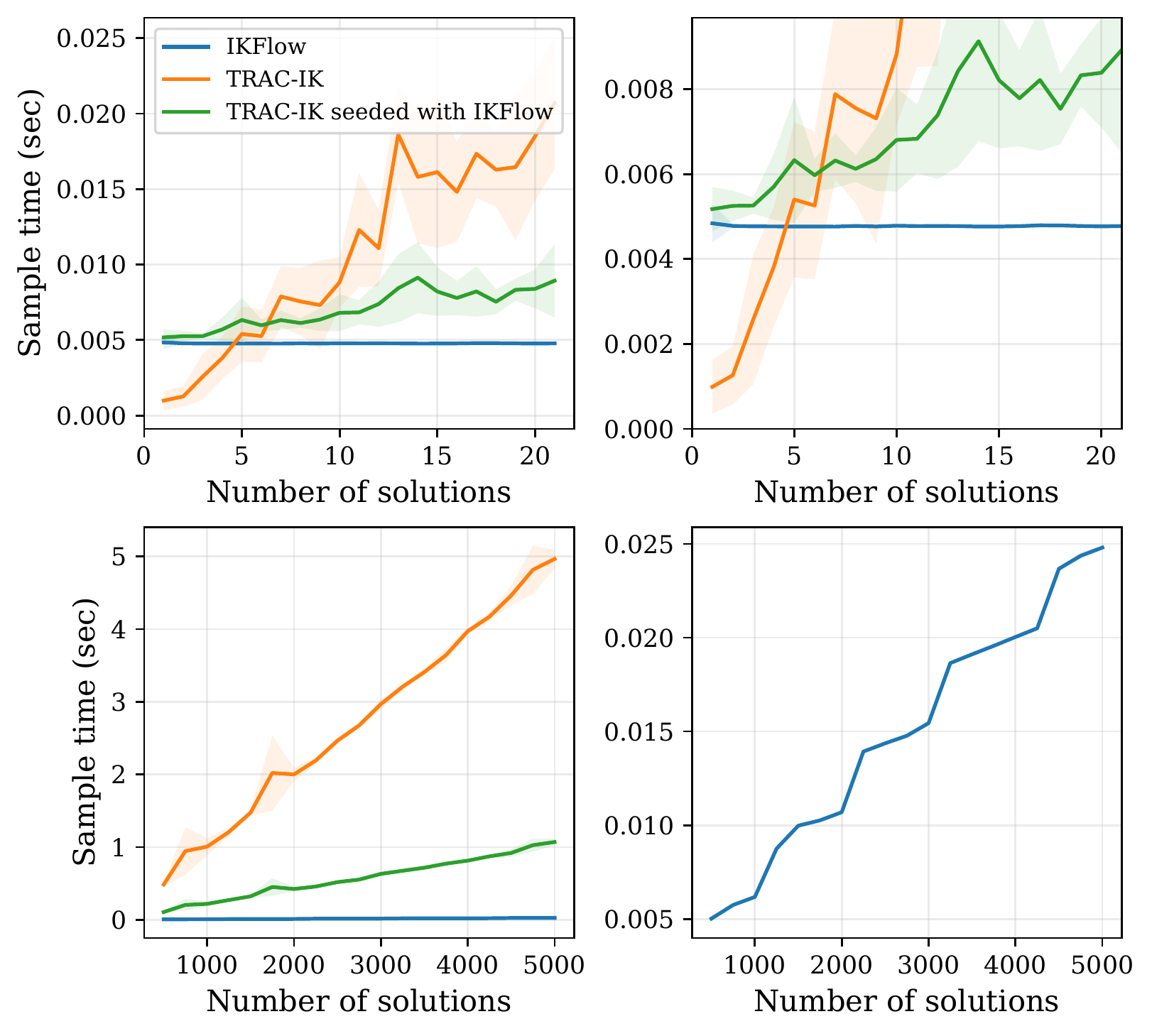}
    \caption{The run time of IKFlow, TRAC-IK, and TRAC-IK when seeded with solutions returned by IKFlow as a function of the number of requested solutions for the ATLAS Arm. The relationship for the IKFlow model is roughly linear. The steps in the IKFlow graph correspond to the size of the batch that can be fit on the GPU. Seeding TRAC-IK with IKFlow provides the 1e-6 accuracy of TRAC-IK with one fifth the runtime of TRAC-IK.}
    \label{fig:runtime_scale}
\end{figure}
%

\section{Results}
Our results demonstrate that IKFlow provides a representative set of solutions, quickly, with acceptable error. The results of all the experiments can be found in Table~\ref{table:results}. 

\subsection{Comparative Evaluation}
Benchmarking results are shown in Figure~\ref{fig:benchmarking}. On both test beds, the IKFlow model performs considerably better than the other two models. These results are consistent with previous findings~\cite{Kruse2019BenchmarkingProblems}, in which it is shown that Conditional INNs (cINNs) outperform INNs on the railY\textunderscore chain3 testbed. The Panda Arm test bed demonstrates that IKFlow handles higher dimensional problems better than the INN and MDN. The higher the dimension of the problem, the more likely there is to be a lower dimension solution space that would introduce instabilities in training.

\subsection{Accuracy}
The accuracy of the system output ranges from 7.72mm to 0.36mm and from 2.81 degrees to 0.15 degrees. For a point of reference, the mechanical repeatability of many industrial arms is 0.1mm. This level of accuracy is sufficient for many tasks; additionally these solutions can also be quickly refined with numerical optimization approaches to reach arbitrary levels of precision. For ATLAS - Arm, refinement takes on average 0.20 ms, as presented in Figure~\ref{fig:runtime_scale}.

%
\begin{figure}
    \centering
    \includegraphics[scale=0.4]{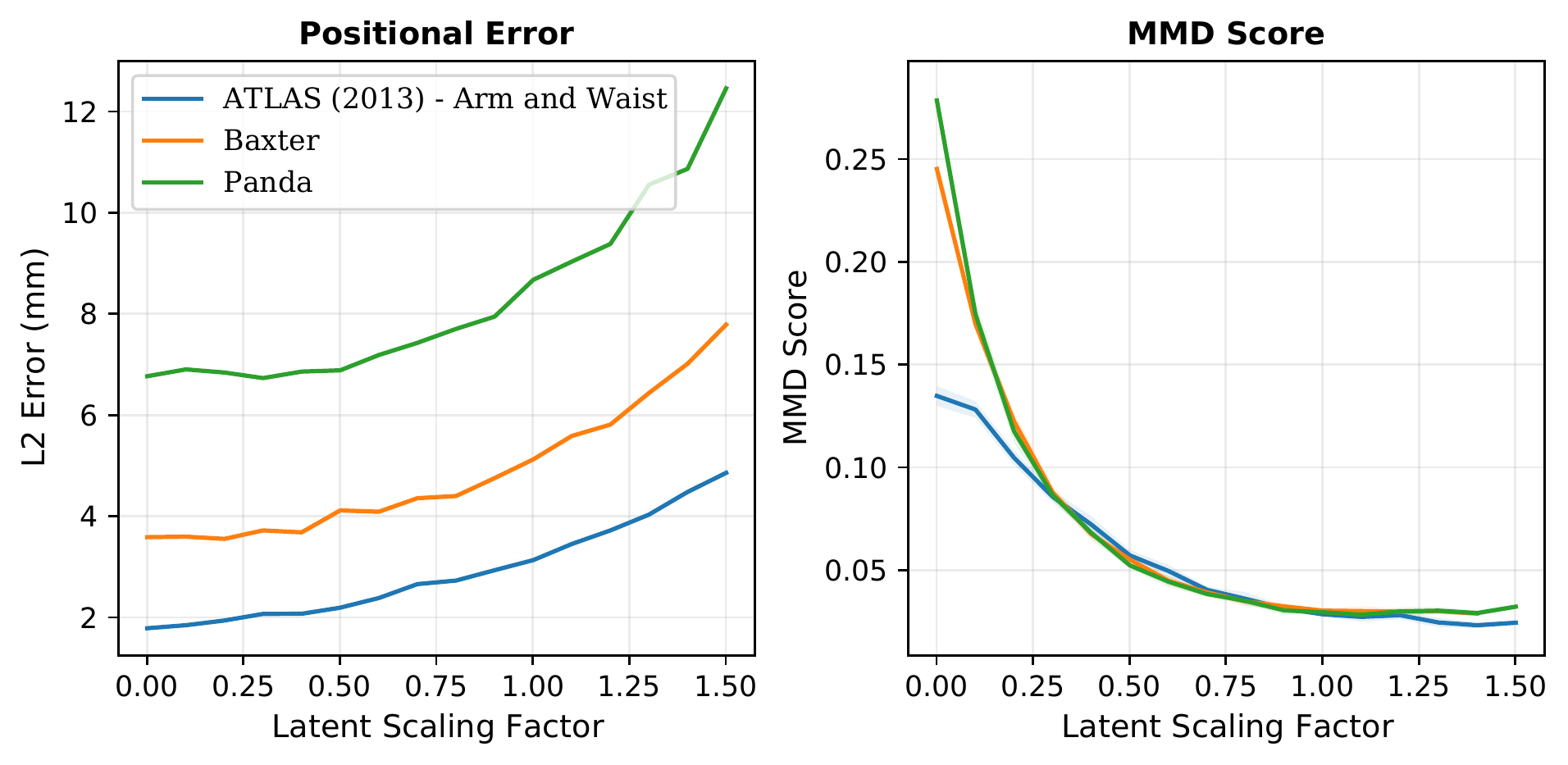}
    \caption{Positional Error and MMD Score of IKFlow as a function of the scaling factor of the latent noise used to sample from the model. Reducing the scaling value increases the models accuracy at the cost of lowering the diversity of the returned solutions.}
    \label{fig:latent_scaling}
\end{figure}
%

\subsection{Runtime}
The runtime of the approach is fast enough to enable its use as a sub-routine in other algorithms. Nonlinear optimization approaches find a single solution in about 0.3 millisecond~\cite{Beeson2015TRAC-IK:Kinematics}, whereas IKFlow can return 500 solutions in 5ms. Figure~\ref{fig:runtime_scale} also demonstrates that the approach scales linearly with the number of requested solution samples. The gradient of the increase is low with $4,000$ samples found in $20$ milliseconds. This means that even complex solution sets can be approximated quickly. For a point of comparison, if TRAC-IK, a common nonlinear optimization based IK solver ~\cite{Beeson2015TRAC-IK:Kinematics}, is fed random seeds in hopes that the local minima it finds are different, it takes more than a second to return $1,000$ samples. 

Notably however, TRAC-IK takes approximately 5x less time to run when seeded with an approximate solution returned by IKFlow. Empirically, it is faster to first run IKFlow to generate seeds before running TRAC-IK when requesting 7 or more solutions.

\subsection{Solution Space Coverage}
Figure~\ref{fig:comparisons} provides a qualitative comparison of ground truth samples with samples from IKFlow. The solution spaces look quite similar, and provide reference points for the MMD score for other chains. All of the kinematic chains included have a MMD score under $0.05$. This implies that the IKFlow solution spaces are very similar to the ground truth samples, with only a few erroneous solutions, or small gaps in coverage. 


\subsection{Limitations}

\begin{figure}
    \centering
    \includegraphics[scale=0.35]{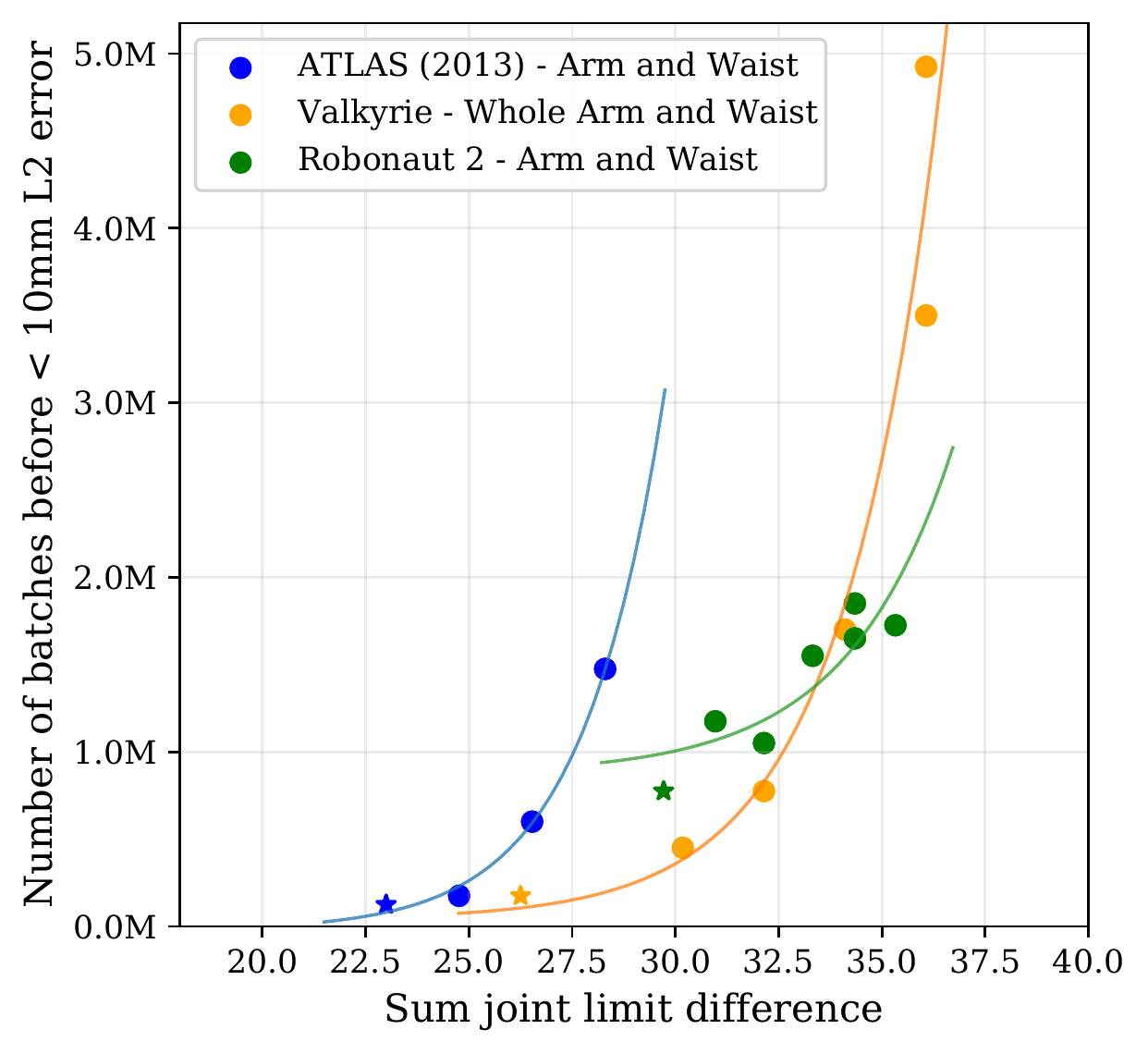}
    \caption{The number of training batches before the IKFlow model reaches an average of 1cm L2 position error as a function of the sum of the joint limit ranges of the kinematic chain. The actual robot is plotted as a star - the other kinematic chains have artificially expanded joint limits. The results indicate that the number of training batches to reach 1cm of error grows exponentially with the sum of the joint limit ranges.}
    \label{fig:SJR}
\end{figure}

While the proposed method is shown to accurately model kinematic chains with L2 Position error as low as 0.36 mm, it has been found that the training time required for models to reach a given position error grows with the complexity of the kinematic chain that is being modeled. While there exist geometrically derived formulations of kinematic complexity, we choose a simple formulation---the sum of the differences between the upper and lower limits for every joint in the robot:
\begin{equation}
\label{eq:sjld}
\sum_{i=1}^{n} u_i - l_i,
\end{equation}
where $u_i$ and $l_i$ are the upper and lower limits, respectively, for joint $i$. Here this is referred to as the sum of joint limit ranges. An experiment was performed by artificially expanding the joint limits for three robots and measuring the resulting number of training batches required for the respective IKFlow models to reach 1cm of error. The results indicate that the number of training batches required for IKFlow to reach a given L2 error grows exponentially with an increase in equation~\ref{eq:sjld} of the kinematic system it is modeling. The results of this experiment are shown in Figure~\ref{fig:SJR}.


%
%

\section{Conclusion}
IKFlow is a novel IK solver capable of providing quick, accurate, and diverse solutions for kinematically redundant robots operating in $SE(3)$, based on modeling IK solutions as a distribution over joint poses and using deep generative modeling to model these distributions. Our experiments demonstrated that IKFlow can generate hundreds to thousands of solutions covering the solution space in milliseconds. The average pose error of the results showed that our approach is capable of finding solutions with millimeters of translation error, and less than 1.5 degrees of rotational error. These results demonstrate that IKFlow can serve as the basis for expanded functionality of 7+ DOF kinematic chains.




\section*{APPENDIX}

\subsection{Model and Training Parameters}
The coefficient networks are 3x1024-wide fully connected networks with Leaky-Relu activation. A softflow noise scale of $1 \times 10^{-3}$ was used across all of the models. The learning rate was set to $5 \times 10^{-4}$ and decayed exponentially by a factor of $0.979$ after every $39000$ batches. Models were trained until convergence on 2.5 million points using the Ranger optimizer with batch size 128 using a NVIDIA GeForce RTX 2080 Ti graphics card.

\subsection{Benchmarking Implementation}
For the railY\textunderscore chain3 test, the IKFlow and INN models both have 6 coupling layers with 3x1024-wide fully connected networks and a latent space dimension of 5. For Panda Arm, both models have 12 coupling layers with 3x1024-wide fully connected coefficient networks and a latent space dimension of 10. The Mixture Density Network (MDN) has 70 and 125 mixture components for the railY\textunderscore chain3 and Panda arm respectively. MDN models were implemented using the \textit{tonyduan/mdn} repository~\cite{DuanGitHubPyTorch.}. 



\section*{ACKNOWLEDGMENTS}
This research was supported in part by DARPA under
agreement number D15AP00104, and the ONR under the PERISCOPE MURI Contract N00014-17-1-2699. The content is solely the responsibility of the authors and does not necessarily represent the official views of DARPA. Barrett Ames was partially supported by a NDSEG Fellowship. The authors would like to thank Ben Burchfiel, and Colin Devine for their support of this work. 

\Urlmuskip=0mu plus 1mu\relax
\bibliographystyle{plainnat}

\bibliography{references.bib}

\end{document}